\documentclass[11pt]{article}
\usepackage[]{acl}
\usepackage{times}
\usepackage{latexsym}

\usepackage[T1]{fontenc}

\usepackage[utf8]{inputenc}

\usepackage{microtype}

\usepackage{inconsolata}

\usepackage{graphicx}

\usepackage{microtype}
\usepackage{graphicx}
\usepackage{booktabs} 
\usepackage{multirow}
\usepackage{booktabs}
\usepackage{subcaption}
\usepackage{tabularx}
\usepackage{xcolor}         
\usepackage{colortbl}
\usepackage{nicefrac} 

\usepackage{amsmath,amssymb}
\usepackage{paralist}

\definecolor{MyDarkGreen}{rgb}{0.02,0.6,0.02}
\definecolor{Red}{rgb}{0.7,0.01,0.01}

\usepackage{hyperref}

\newcommand{\tao}[1]{}
\newcommand{\kw}[1]{}
\newcommand{\yxy}{(y_i|\mathbf{x,y}_{<i})}
\newcommand{\xyi}{(\mathbf{x,y}_{<i})}
\newcommand{\xyic}{(\mathbf{x,y}_{<i}\oplus y_i)}
\newcommand{\byx}{(\mathbf{y|x})}
\newcommand{\bxy}{(\mathbf{x,y})}

\title{Attribute Controlled Fine-tuning for Large Language Models: A Case Study on Detoxification}

\author{
Tao Meng$^{1,2}$, Ninareh Mehrabi$^{2}$, Palash Goyal$^{2}$, Anil Ramakrishna$^{2}$, Aram Galstyan$^{2}$ \\
\bfseries  Richard Zemel$^{2}$, Kai-Wei Chang$^{1,2}$, Rahul Gupta$^{2}$, Charith Peris$^{2}$ \\
$^1$ University of California, Los Angeles \\
$^2$ Amazon.com, Inc. \\
\texttt{\{tmeng, kwchang\}@cs.ucla.edu, perisc@amazon.com} \\
  }
\begin{document}
\maketitle

\begin{abstract}
We propose a constraint learning schema for fine-tuning Large Language Models (LLMs) with attribute control. Given a training corpus and control criteria formulated as a sequence-level constraint on model outputs, our method 
fine-tunes the LLM on the training corpus while enhancing constraint satisfaction with minimal impact on its utility and generation quality. 
Specifically, our approach 
regularizes the LLM training by penalizing the KL divergence between the desired output distribution, which satisfies the constraints, and the LLM's posterior. This regularization term can be approximated by an auxiliary model trained to decompose the sequence-level constraints into token-level guidance, allowing the term to be measured by a closed-form formulation. To further improve efficiency, we design a parallel scheme for concurrently updating both the LLM and the auxiliary model. 
We evaluate the empirical performance of our approach by controlling the toxicity when training an LLM. We show that our approach leads to an LLM that produces fewer inappropriate responses while achieving competitive performance on benchmarks and a toxicity detection task. 

\end{abstract}

\section{Introduction}
\label{sec:intro}
    Large language models (LLMs) have demonstrated impressive performance across a variety of tasks which has led to their widespread adoption for a multitude of AI applications. However, they carry the risk of producing inappropriate, unsafe, unfair outputs~\citep{wallace2019universal, sheng2019woman, gehman2020rtp,huang2024trustllm}
    Ideally, LLMs should learn to comply with constraints and policies specified by users. 
    For example, in a user-facing application like a chatbot, LLMs should never generate toxic or offensive responses, nor to divulge sensitive information. While there are several post hoc methods to moderate LLM outputs~\cite{lu2022neuro, qian2022controllable,markov2023holistic}, they lack an efficient and principled approach to training LLMs to adhere to constraints. 
   

    We start by defining a sequence-level oracle as a function that takes an LLM's output and adjudicates whether it satisfies a predefined set of attribute constraints. In practice, the oracle can be a rule-based, model-based, or mixed system (e.g., a classifier that decides whether a sentence is toxic). Given a pre-trained LLM and the oracle, we aim to fine-tune an LLM to achieve the following: 1) \textbf{Attribute control:} The LLM output passes the oracle with a high probability. 2) \textbf{Utility preservation:} The LLM maintains performance comparable to the original LLM on utility benchmarks. 3) \textbf{Training efficiency:} The cost of fine-tuning with attribute control is similar to that of the typical fine-tuning.


    While existing approaches can meet some of these criteria, achieving all of them is challenging. For example, filtering training data with the oracle function before fine-tuning~\citep{wang2022exploring} is a simple and efficient method.    
    However, this approach could be less effective. Taking toxicity control as an example, if we filter out the toxic data from a fine-tuning corpus, in a regular context the model will learn not to generate toxic contents. Nevertheless, it might still be possible to trigger the generation of offensive responses given a toxic prompts, due to the fact that toxic prompts are out-of-distribution in relation to the fine-tuning corpus.
    Another promising approach is reinforcement learning (RL) considering controlling criteria in the reward function~\citep{snell2023offline, mudgal2023controlled}.
    However, RL setups tend to be inefficient and require preference data generation which adds significant overhead in comparison to generic fine-tuning.

    In this work, we propose a novel solution to training an LLM with a set of attribute constraints. Inspired by the classic idea of constraint-driven learning~\citep{chang2007guiding} and posterior regularization \citep{ganchev2010posterior}, we incorporate constraints as a regularizer in fine-tuning. 
    Specifically, we estimate the closest distribution from the current model that satisfies the constraints and penalize the gap from the current model distribution to this estimated distribution to regularize the LLM during fine-tuning. We iterate through this process to push the LLM closer to the feasible region of generations, making the estimation progressively more accurate. 

    This iterative fine-tuning process updates the base LLM and regularizer sequentially, causing run time to be significantly longer than the typical fine-tuning. Thus, we parallelize our algorithm by updating the base LLM and regularizer simultaneously based on their status in the last iteration. Empirically, the parallelization achieves the same level of performance compared to sequential fine-tuning, and the time complexity is the same as a typical fine-tuning approach.

    To validate the effectiveness of our proposed method, we conduct a case study in detoxification, considering three scenarios involving different datasets. In the first scenario, we fine-tune LLMs on datasets rich in toxic language with an attribute control that prevents the generation of toxic outputs. Our approach successfully passes stress tests and produces responses with lower toxicity compared to all baseline models. In the second scenario, we explore whether the attribute control can retain the utility of the LLM while reducing the toxicity of its responses. Training only on a small dataset will lead to catastrophic forgetting. Therefore, we fine-tune the LLM on a mix of data comprising toxiGen~\citep{hartvigsen2022toxigen} and Wikitext~\citep{merity2016wikitext} datasets with attribute control. Our method demonstrates the best balance between model utility and toxicity management compared to similar techniques.
    
    Finally, we assess whether the LLM can effectively identify toxic content without generating it, a critical skill since the model must recognize toxic elements to avoid producing them. In standard fine-tuning, these goals often conflict: the model learns to identify toxicity through training on a toxic corpus, which paradoxically increases the generation of toxic content. However, our method successfully mitigates the generation of toxic content while maintaining classification performance on par with traditional fine-tuning techniques.



    We summarize our {\bf contributions} as follows: 
    \begin{itemize}
        \item We provide an efficient and effective solution to the attribute-controlled fine-tuning. 
        \item Empirically, we achieve the current best trade-off between attribute control (measured using toxicity) and utility performance against a suite of baselines. 
        \item We show that our approach enables the model to retain knowledge of the concept of a given attribute 
        and yet selectively choose to avoid generating it. This can not be achieved via generic fine-tuning. 
    \end{itemize}
    



\section{Related Work}

    Prior work exists on the controlled generation problem and it can be divided into two fronts. Solutions that apply at inference time during decoding, and solutions that apply at fine-tuning. 

    \paragraph{Attribute controlled decoding for LLMs}
    Several methods have been explored to control LLM generation during decoding. Some prominent methods include activation editing~\citep{hernandez2023measuring,li2023inference} which adjusts the activation vectors in the LLM, and weight editing~\citep{meng2022locating, ilharco2023editing} which adjusts the weights in LLM. \citet{dathathri2020pplm} (PPLM) leverages an auxiliary model to steer the base LLM distribution. Following this line of work, \citet{krause2021gedi} (GeDi) and \citet{liu2021dexpert} (DExpert) used contrastive learning as an objective to achieve attribute control during decoding. \citet{yang2021fudge} (FUDGE) leveraged an external token-level auxiliary model for their work. This was followed by \citep{meng2022nado} (NADO) who trained a token-level auxiliary model by decomposing the controlling criteria via optimization and approximation. \citet{zhang2023gelato} (GeLaTo) leverage a probabilistic circuit to tractably incorporate symbolic constraints. 

    Our work is inspired by NADO~\citep{meng2022nado}, and we take it as a sub-component in our fine-tuning approach. However, our paper significantly differs from NADO. Firstly, NADO presents an inference method that guides generation by reweighting output distributions without updating the model weights of the base model. In contrast, our approach involves fine-tuning the model. We use NADO to estimate the optimal distribution that satisfies the constraints. However, our design of the posterior regularizer, the iterative-updated scheme, and the parallel computing algorithm is novel. Moreover, the effectiveness of regularizing model training with constraints has not been studied for LLMs. We also demonstrate our approach in a real-world application, enhancing models to understand toxicity while preventing the generation of toxic content. This cannot be done by NADO as their approach is only a decoding method.
    
    \paragraph{Attribute controlled fine-tuning for LLMs}
    When controlling attributes during fine-tuning, the most straightforward way is to filter out training data that contain the attribute~\citep{wang2022exploring}. However, this approach tends to cause performance degradation as it can filter out large portions of the training set and does not actively leverage negative examples. Neuro-symbolic approach~\citep{kareem2023pseudo} incorporates symbolic constraints as loss added to the training objective; however, it cannot handle complex or implicit constraints. RL~\citep{ramamurthy2023rl,snell2023offline,mudgal2023controlled} can be utilized to control an attribute via the use of an attribute-related reward. RLHF~\citep{ouyang2022training, xu2022learning,ziegler2019fine,bai2022training} and RLAIF~\citep{bai2022constitutional,lee2023rlaif} leverage feedback from humans and LLMs, respectively, to control the required attributes. However, they are more focused on being aligned with humans (or LLMs) rather than specific attribute control. RL-based methods are effective but often inefficient due to the large variance in feedback provided by the reward models. 

    \kw{I feel we should discuss some earlier work on constraint-driven learning and posteior regulaurzation and argue them are mostly apply in either linear models or probabilistic graphic models. And discuss they usually based on strong independent assumptions. } 

\section{Methodology}
    

        \begin{figure}[t]
            \centering
            \includegraphics[trim={4cm 0 0 0}, width=0.3\linewidth]{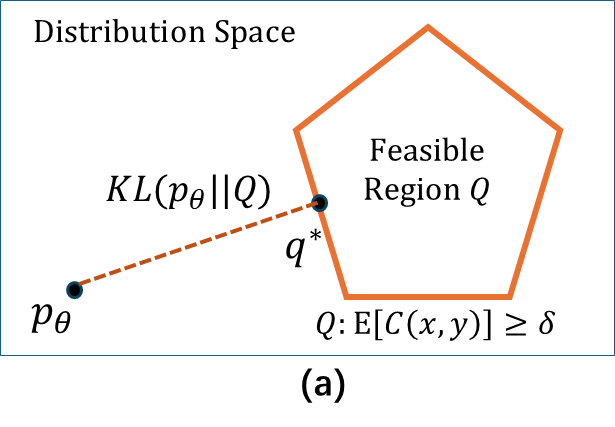}
            \includegraphics[trim={0 0 4cm 0}, width=0.3\linewidth]{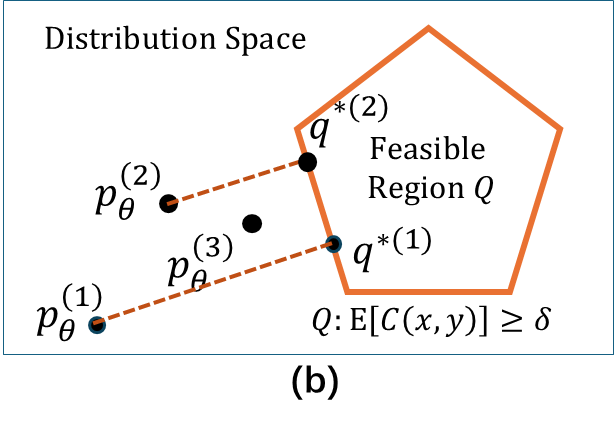}
            \caption{A conceptually visualization of base LLM distribution $p_\theta$ and optimal distribution $q^*$ in fine-tuning. The polygon is representing the feasible region $Q$ where the constraints are satisfied. On (a) it shows the regularizer term is defined as the closest distance from $p_\theta$ to $Q$. Regularized by KL-divergence from $q$, on (b) we show the LLM distribution $p_\theta$ is gradually pushed towards the feasible region.}
            \label{fig:push}
        \end{figure}
    \subsection{Notation and Formalization}
    \label{sec:overview}
        We use $p_\theta$ to denote the LLM and $\theta$ is its trainable weights, $\mathbf{x} \in \mathcal{X}$ is the input (e.g. prompts), and $\mathbf{y}$ is the generated output sequence of the model. We denote $\mathbf{y}_{<i}=(y_0, y_1,\dots, y_{i-1})$ as the prefix of $\mathbf{y}$. $C\bxy:\mathcal{X}\times\mathcal{Y}\rightarrow \{0,1\}$ denotes a black box oracle function which takes prompt $\mathbf{x}$ and model output $\mathbf{y}$ as input, and outputs whether the generation $\mathbf{y}$ satisfies the constraints.\footnote{We can extend our approach to handle real value constraints in the form of  $C\bxy \in [0,1]$. For simplicity, we consider binary constraints  in this paper.}
        For example, in detoxification, the oracle takes a user prompt $\mathbf{x}$ and the model response $\mathbf{y}$ as input, then returns $0$ when the response is offensive, indicating that the response is unacceptable. 

        Given an LLM $p_\theta$, a black box oracle $C\bxy$, and a training dataset $D=\{\mathbf{x}_i,\mathbf{y}_i\}_{i=1}^N,$ our goal is to fine-tune the LLM as $ p_{\tilde\theta}$ so that 
        the model retains its utilities while 
        satisfying the constraints in expectation:
        \begin{equation}
        \label{eq:constr}
            \forall \mathbf{x}\in \mathcal{X},\ \mathbb{E}_{\mathbf{y}\sim p_{\tilde\theta}\byx}[C\bxy]\geq\delta.
        \end{equation}
        Here, $\delta$ is a user-specified parameter. When $\delta=1$, the training will push the model to satisfy all constraints, while choosing $0<\delta<1$, soft constraints are enforced.  



    \subsection{Fine-tuning LLM with Posterior Regularization}
    \label{sec:finetune}
        Given a training data $\bxy,$ typically the objective we fine-tune the LLM $p_\theta$ is defined as
        $$L_{LM}(p_\theta;\mathbf{x,y})=\sum\nolimits_i L_{CE}(p_\theta\yxy,1),$$
        where $L_{CE}$ is the cross-entropy loss. To achieve attribute control, we propose to add a regularization term that penalizes the violation of constraints. 

        The general idea of our approach is to fine-tune LM with a regularizer to penalize the 
        following posterior regularization~\citep{ganchev2010posterior}, we define
        \begin{equation*}
        \begin{aligned}
            &Q:=\{q\ |\ \forall \mathbf{x}\in\mathcal{X}, \mathbb{E}_{\mathbf{y}\sim q\byx}[C\bxy]\geq\delta\}\\
            &D_{KL}(p_\theta\|Q):=\min\nolimits_{q\in Q} D_{KL}(p_\theta\|q).
        \end{aligned}
        \end{equation*}
        The feasible region $Q$ is the set of distributions that satisfy the constraint in Eq.~\eqref{eq:constr}. Illustrated by Fig.~\ref{fig:push}(a), the regularization term $D_{KL}$ is defined as the smallest divergence from $Q$ measured by Kullback–Leibler (KL) divergence. The overall objective of fine-tuning is
        \begin{equation}
        \label{eq:pr}
            L(p_\theta;\mathbf{x,y},Q):=L_{LM}(p_\theta;\mathbf{x,y})+\lambda D_{KL}(p_\theta\|Q),
        \end{equation}
        where $\lambda$ is the hyper-parameter balancing the two terms.


        However, the second term is intractable and hard to compute: when the base model distribution changes in fine-tuning, the closest distribution also changes. To address this issue, we design an iterative fine-tuning process: we first fix the base model distribution and estimate the closest distribution in the feasible set (Sec. \ref{sec:opt}, \ref{sec:aux}), and then we fix the estimated distribution as the reference distribution in the KL regularizer to fine-tune the LLM (Sec. \ref{sec:seq_ft}). To speed up the process, we further propose parallel fine-tuning (Sec. \ref{sec:parallel}).
        
        \kw{It might be hard to understand this. Now It think you may add this overview in Sec 3.2 after defining Q, but we would need to discuss it in pure English to motivate the algorithm without those notations.  }\tao{How about this one}
    \subsection{Optimal Distribution in Feasible Region}
    \label{sec:opt}

        To compute the regularizer term in fine-tuning, we need to find the optimal distribution $q^*$ as the reference distribution by solving the following problem
        \begin{equation}
        \label{eq:opt}
            q^*=\arg\min\nolimits_{q:E_{\mathbf{y}\sim q(\mathbf{y|x})}[C\bxy]\geq\delta} D_{KL}(q\|p).
        \end{equation}
    

        \citet{meng2022nado} shows the close-form solution can be derived as
        \begin{equation*}
            \begin{aligned}
                &q^*\yxy\propto p_\theta\yxy\cdot \\
                & [(\delta-R_C^p(\mathbf{x}))R_C^p \xyic +(1-\delta)R_C^p(\mathbf{x})]
            \end{aligned}
        \end{equation*}
        if $\delta>R^p_C(\mathbf{x})$. Otherwise the constraint is already satisfied and $q^*\yxy = p_\theta\yxy.$
        
        Specifically, when $\delta=1$, we have
        \begin{equation}
        \label{eq:qstar}
            q^*\yxy\propto p_\theta\yxy R_C^p \xyic,
        \end{equation}
        where $\oplus$ is the concatenation operation. $R_C^p\xyi$ is the probability that the generated output will satisfy constraints when the generation finishes given input $\mathbf{x}$ and prefix $\mathbf{y}_{<i}$, and is given by  
        \begin{equation}
        \label{eq:rc}
            \begin{aligned}
                &R^p_C\xyi=\Pr\nolimits_{\mathbf{y}\sim p_\theta(\mathbf{y}|\mathbf{x,y}_{<i})}[C\bxy=1],\\
                &R^p_C(\mathbf{x})=\Pr\nolimits_{\mathbf{y}\sim p_\theta(\mathbf{y}|\mathbf{x})}[C\bxy=1].
                \nonumber
            \end{aligned}
        \end{equation}

        Basically, the satisfaction probability $R^p_C$ is the token-level decomposition of the sentence-level oracle $C$. Based on $\delta$ and $R^p_C,$ the solution shows how to adjust the next token distribution from the original distribution $p_\theta.$
        \kw{can you add a sentence or two to provide the intuition behind this solution. } \tao{Done.}

        Unfortunately, although the function $R_C^p$ is well-defined, it is not tractable. To achieve the optimal solution in Eq.~\eqref{eq:opt}, in this work, we estimate $R_C^p$ from the training data and the LLM, and update the two terms in Eq.~\eqref{eq:opt} iteratively. In sections~\ref{sec:aux} and~\ref{sec:seq_ft}, we describe how we estimate $R_C^p$ from the data and the current model $p_\theta$, and how we update the model $p_\theta$ with the help of the estimated $R_C^p$.

        Note that in fine-tuning objective Eq. \eqref{eq:pr}, the reference distribution $q$ is fixed, and we update $p_\theta,$ so the regularizer is $D_{KL}(p_\theta\|q)$. However, here the model $p$ is fixed and we seek the optimal $q$, so we minimize $D_{KL}(q\|p_\theta).$ Empirically, when we optimize the KL divergence term, the trainable weights in the reference distribution usually lead to unstable training. Thus, we always set the fixed distribution as the reference distribution.
        \kw{Can you say a bit more and motivate why the order has to be set in this way.}\tao{Done.}

    \subsection{Estimating $R_C^p$ from LLM and Data}
    \label{sec:aux}
        To estimate $R_C^p$, we train an auxiliary model $R_\phi$ from the training data $\tilde{D}$ weighted by the base LLM $p_\theta.$ 
        We assume the empirical distribution is drawn from unseen training distribution $D$, and has no repetition\footnote{If there are repeated examples we can remove them before training. This assumption makes sure that the following weighted empirical loss mimics the expectation loss from sampling.},
        \kw{this description is a bit weird. Can we say we assume there is no repeated examples in $D$ and put a footnote say if there are repeated examples we can remove them before the training. Also, you may want to motivate why you need this assumption.}\tao{Done.}
        and set the objective function for a particular example $(\mathbf{x,y})\in\tilde{D}$ as the cross-entropy loss between the predicted satisfaction probability and oracle output, weighted by the sequence probability $p_\theta\byx$
        \begin{equation}
        \label{eq:weight}
            \begin{aligned} &L(R_\phi;\mathbf{x,y}) \\ = &p_\theta\byx\sum\nolimits_i L_{CE}(R_\phi\xyi, C\xyi).
            \end{aligned}
        \end{equation}
        \kw{This objective is define for a particular example $(x,y)\in D$? }\tao{Yes, description fixed.}
        
        Considering the expected loss on distribution $D$, 
        we have
        \begin{equation}
        \small
        \label{eq:loss}
            \begin{aligned} &\mathbb{E}_{\tilde{D}\sim D, \bxy\sim\tilde{D}}[L(R_\phi;\mathbf{x,y})]\\
        =&\mathbb{E}\!_{\tilde{D}\!\sim\! D, \!\bxy\sim \tilde{D}}[p_\theta(\mathbf{y|x})\!\sum_i L_{CE}\!(\!R_\phi\xyi, C\bxy\!)\!]\\
        =&\mathbb{E}_{\mathbf{x}\sim D, \mathbf{y}\sim p_\theta(\mathbf{y|x})}[\sum_i L_{CE}(R_\phi\xyi, C\bxy)]\\
                =&\sum_i L_{CE}(R_\phi\xyi, R_C^p\xyi).
            \end{aligned}
        \end{equation}
        Therefore, the global minimum of the expected loss function is reached when $R_\phi\xyi=R_C^p\xyi$.
        \kw{I'm not sure which objective function you refer to, also you may need to explain why the local minimum is that}\tao{how about this}

        In \citet{meng2022nado} the auxiliary model is trained by the data sampled from $p_\theta$ without weighting the data by its probability as Eq. \eqref{eq:weight}. The expected loss is the same as Eq. \eqref{eq:loss}. \kw{it may not clear what do you mean by "weighting the data"}
        \kw{The same as what?}\tao{is it clear now}
        In our experiments, we apply sampling to train the auxiliary model, when there is no available training data. Hereafter in this work, we follow \citet{meng2022nado} and refer to this auxiliary model as the neurally-decomposed oracle (NADO). In practice, NADO architecture is similar as the base LLM, with the same hidden dimension and fewer layers.
        \kw{I feel here you can add a few more details. You can mention that in practice, we can use a smaller model as NADO.  }\tao{Added some details}

    \subsection{Iteratively Updating $p_\theta$ by Regularized Fine-tuning}
    \label{sec:seq_ft}
        Once we estimate $R^C_p$ by NADO $R_\phi,$ we are able to get the estimated optimal distribution $q$ from Eq.~\eqref{eq:opt} by replacing $R^C_p$ with $R_\phi$. We then plug in the estimated optimal distribution to the fine-tuning objective in Eq.~\eqref{eq:pr} as
        \begin{equation}
        \label{eq:obj}
            \begin{aligned}
                &L(p_\theta;\mathbf{x,y},q)\\
                =&L_{LM}(p_\theta;\mathbf{x,y})+\lambda D_{KL}(p_\theta(\mathbf{y|x})\|q(\mathbf{y|x})) \\
                =&\sum\nolimits_i \log p_\theta\yxy \\
                +& \lambda D_{KL}(p_\theta\yxy\|q\yxy).
            \end{aligned}       
        \end{equation}
        \kw{$q(\mathbf{y|x}$ is not defined? Also $q\yxy$ is the same as $q*\yxy$ in Eq (4)? It's okay to not add * here bu then in that case we need to explain. }\tao{$q^*$ refers to the close-form optimal distribution (intractable), and $q$ is the estimated one (by replacing $R^C_p$ with $R_\phi$.}


         \begin{figure*}
            \centering
            \includegraphics[trim={4cm 0 0 0}, width=0.4\linewidth]{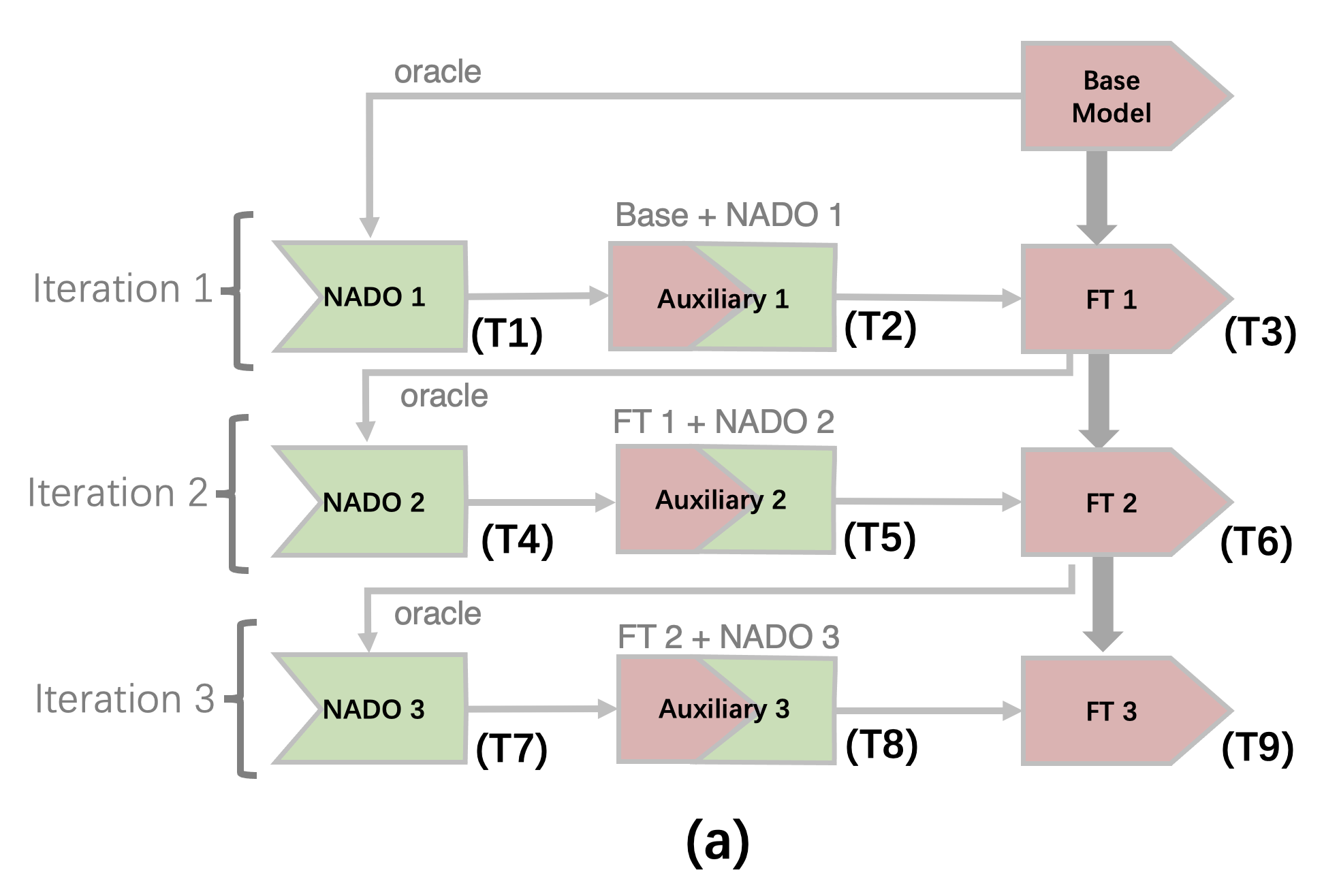}
            \includegraphics[trim={0 0 4cm 0}, width=0.4\linewidth]{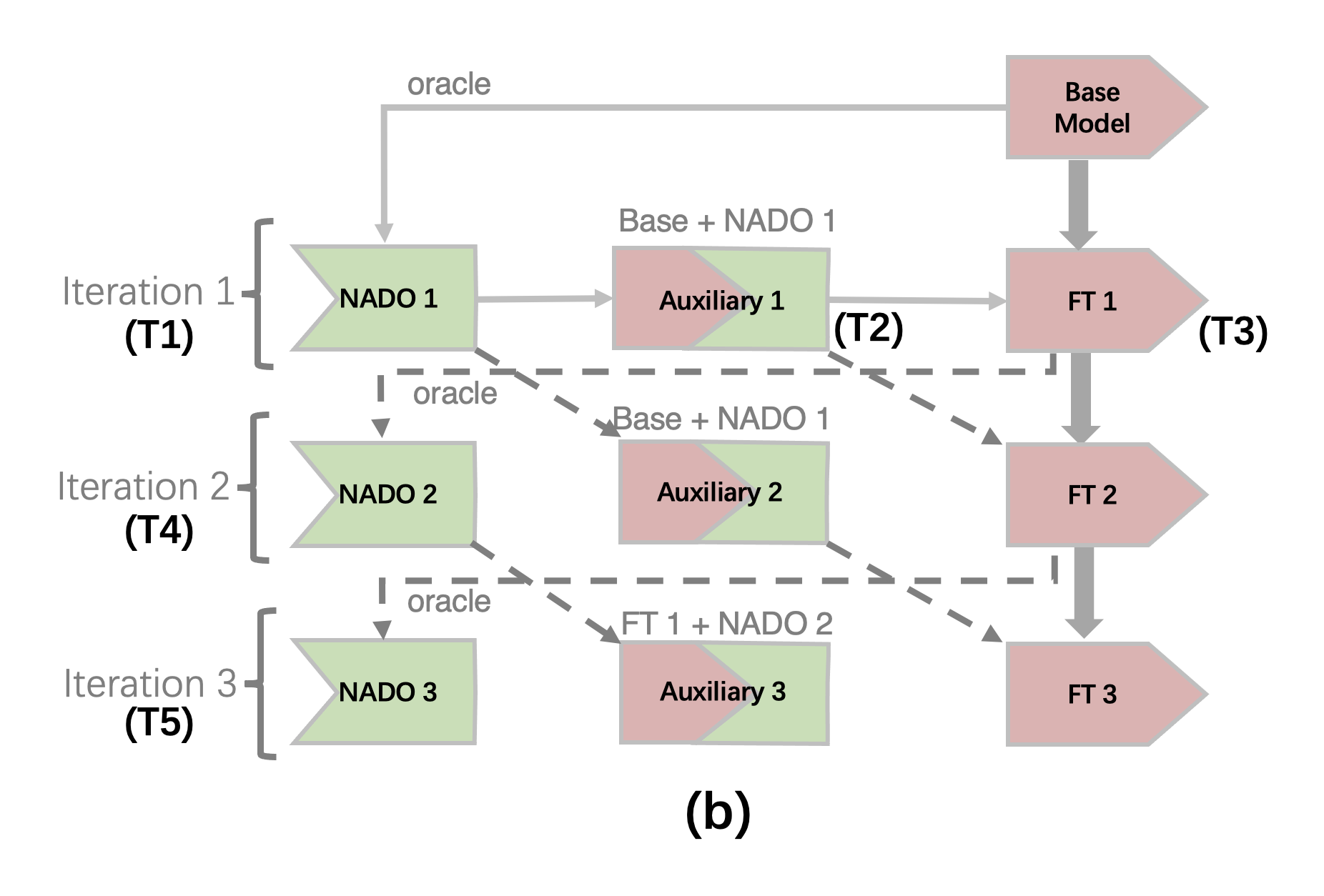}
            \caption{An illustration of sequential and parallel fine-tuning for three iterations. We use $T$ (time step) to indicate the time. Oracle, symbolizes the process of sampling data from an LLM, labeling with an oracle, and training the NADO model. On the left, we show sequential execution with the grey arrows showing the direction of flow. On the right, we show the parallelized execution. Note that in this case, all components (left to right) of each iteration are run at the same time step (except in iteration 1). Note also, that the grey dashed arrows (from iteration 2 onwards) do not flow across components within the same iteration level, indicating the independence of each component from other components in the same level. This allows them to be executed in parallel.}
            \label{fig:schematic}
        \end{figure*}

        Intuitively, a model fine-tuned with the objective in Eq.~\eqref{eq:obj} exhibits a trade-off between the model quality and the amount of control. Fine-tuned on this objective, the model converges at some mid-point between $p_\theta$ and $q$. 
        
        Now we are able to estimate $R^C_p$ by $R_\phi$ from the training data and model $p_\theta$ (Sec.~\ref{sec:aux}), and fine-tune $p_\theta$ with estimated optimal distribution $q$ derived from $R_\phi$. A straightforward way is to update these models iteratively, which we call ``sequential fine-tuning''. 
        In this process, we gradually push the base model distribution towards the feasible region, and the estimated optimal distribution is more accurate. 
        As shown in Fig.~\ref{fig:schematic}(a) and described in Sec. \ref{sec:overview}, we iteratively run the following three steps:
        \begin{compactitem}
            \item Based on current LLM $p_{\theta}^{(i)}$, sample or weight data $D^{(i)}$ labeled by the oracle.
            \item Train NADO $R_\phi^{(i)}$ using the data $D^{(i)}$ initialized with $R_\phi^{(i-1)}$.
            \item Fine-tune the LLM $p_{\theta}^{(i)}$ with the KL-divergence between $p_\theta^{(i)}$ and $q^{(i)}$ given by Eq.~\eqref{eq:qstar}.
        \end{compactitem}

         The distribution of the base model can be conceptually visualized in Fig.~\ref{fig:push}(b) during fine-tuning. As the base model $p_{\theta}^{(i)}$ getting closer to the feasible region, the estimated optimal distribution $q^{(i)}$ will be more accurate compared to the estimation from the original base model distribution $q^{(1)}$.

    \subsection{Parallel Fine-tuning}
    \label{sec:parallel}
        The iterative fine-tuning process outlined in Section~\ref{sec:seq_ft} executes its steps sequentially by solving the optimization problem in Eq.~\eqref{eq:opt} in each round. However, while accurate, it is also inefficient. In this section, we propose a parallel fine-tuning method to improve efficiency.
        
        In parallel fine-tuning, we propose a set-up that processes the three steps outlined in Sec.~\ref{sec:seq_ft} in parallel (see Fig.~\ref{fig:schematic}(b)). Given $p_{\theta}^{(i)}$, $D^{(i)}$ and $q^{(i)}$, the following three steps are processed simultaneously:
        \begin{compactitem}
            \item Based on current LLM $p_{\theta}^{(i)}$, sample or weight data $D^{(i+1)}$ labeled by the oracle.
            \item Train NADO $R_\phi^{(i+1)}$ using data $D^{(i)}$ initialized with $R_\phi^{(i)}$.
            \item Fine-tune the LLM $p_{\theta}^{(i+1)}$ with the KL-divergence from $q^{(i)}$ given by Eq.~\eqref{eq:qstar}.
        \end{compactitem}
        After one round, we get $p_{\theta}^{(i+1)}$, $D^{(i+1)}$ and $q^{(i+1)}.$ The LLM keeps fine-tuning on the dataset with a regularizer, and the regularizer is updated at every checkpoint. In sequential fine-tuning, the process will terminate at each checkpoint, waiting for the regularizer to update with the data sampled or weighted by the LLM. Compared to a baseline, which fine-tunes without control, the extra time cost in our method is only the extra computation on the regularizer and the time cost in dumping checkpoints. The additional memory cost for NADO is not significant, because it is relatively small compared to the base LLM.

        In practice, we select proper hyperparameters\footnote{Including the number of examples we sample or weight, number of epochs to fine-tune LM, and number of epochs to train NADO.} to ensure the three steps take similar computational time. In such a case, parallel fine-tuning achieves 3x speed up compared to sequential fine-tuning.

    \subsection{Adaptive Regularizer}
        The data for fine-tuning an LLM often includes a diverse mix of sources. Fine-tuning on a specific domain may lead to performance degradation in other domains due to catastrophic forgetting. 
        A popular approach is to add KL-divergence to the original model to avoid the model deviating from the original model~\citep{schulman2017ppo}. To effectively incorporate this mechanism into our approach, we can implement domain-specific regularizers during the fine-tuning process.

        
        Specifically, we denote the training dataset as $D=\bigcup_i D_i$. For each subset $D_i$ with a corresponding constraint oracle $C_i$. We use the base LLM to weight the subset, and $C_i$ to label them. According to Eq.~\eqref{eq:weight}, we train NADO $R_{\phi_i}$ for the constraint oracle $C_i$, and compute the estimated optimal distribution $q_i$. We use $q_i$ as the reference distribution in the KL-divergence. Specifically, when we set $q_i$ as the original distribution, the regularizer sets as the KL-divergence to the original model. We refer to this regularizer as the \textit{preserving regularizer}. 
        \kw{I feel this paragraph is a bit hard to follow, I would suggest first discussing to avoid catastrophic forgetting a popular approach is to add KL divergence to original model to avoid the model devivate from the original model too much. Then you can discuss that in practice, you can add different types of regulairzer. Finally, you can explain what is the setting we consider in the detoxification case. }\tao{Not sure if this is clear.}

        In this work, we demonstrate how to effectively control the toxicity of an LLM while preserving its performance level. We apply the regularize for toxicity control when fine-tuning it on toxicity-related datasets, while using the preserving regularize on a general dataset (like Wikitext).  Formally, we denote $p_0$ as the original LLM, $q$ as the estimated optimal distribution under toxicity constraint oracle, and $D_T\subset D$ as the toxicity-related training set. We adopt the fine-tuning objective in Eq.~\eqref{eq:pr} to
        \begin{equation}
        \label{eq:adaptive}
            \begin{aligned}
                L&(p_\theta;D,q)
                =\sum\nolimits_{\bxy\in D} L_{LM}(p_\theta;\mathbf{x,y})\\
                +&\lambda\sum\nolimits_{\bxy\in D_T} D_{KL}(p_\theta\byx\|q\byx)\\
                +&\lambda\sum\nolimits_{\bxy\notin D_T} D_{KL}(p_\theta\byx\|p_0\byx).
            \end{aligned}
        \end{equation}

\section{Case Study on Detoxification}
    To test the effectiveness of the proposed approach, we apply it to detoxify an LLM. Toxicity, as discussed in Section~\ref{sec:intro}, is of significant importance as a metric for the evaluation of LLM~\citep{brown2020gpt3,touvron2023llama,chowdhery2022palm,touvron2023llama2}. 
    In this context, we apply our fine-tuning schema in three different scenarios; (1) \textbf{detoxification:} testing the effectiveness of our proposed approach in attribute control, (2) \textbf{multi-task scenario:} testing that the controlled model preserves the same level performance on other tasks,   and (3) \textbf{toxicity classification:} testing whether the control affects the model performance on attributes related tasks.  
    
    In all experiments, we set $\delta=1$ to set toxicity as a hard constraint. Detailed notes on data pre-processing, hyper-parameter choice for model training, and the architecture of auxiliary models can be found in the Appendix. 

    
    \subsection{Detoxification }
    \label{sec:detox}
        \begin{table}[t]
            \centering
            \resizebox{0.98\linewidth}{!}{
            \begin{tabular}{c|c|c}
                \toprule
                Model & API Tox. & ToxiGen \\
                \midrule
                Llama baseline &
                0.315 & 23.0 \\
                Reinforcement Learning & 0.269 & 12.3 \\
                NADO Decoding Control & 0.289 & 14.4 \\
                Ours (sequential) & \textbf{0.259} & 11.0 \\
                Ours (parallel) & 0.261 & \textbf{10.9} \\
                \bottomrule
            \end{tabular}}
            \caption{Toxicity scores of Llama-7B model with different detoxification methods. The proposed fine-tuning methods outperform RL and decoding-time control in detoxification. Parallel fine-tuning achieves similar control compared to sequential, with 3x fine-tuning speed.}
        \label{tab:deto}
        \end{table}
        
    Given a corpus and a toxicity oracle, we first show the effectiveness of our approach in detoxification. We also show that parallel fine-tuning achieves a similar performance as the sequential one. \kw{Add a few more sentences to describe the goal/motivation of the experiment and what we hope to see. Also give a brief overview of the setting. }\tao{Done.}
        
    \paragraph{Setup}
    We use Llama-7B~\citep{touvron2023llama} as the base model. NADO has a similar architecture but with only 8 layers. For our experiments, we use RealToxicPrompts (RTP)~\citep{gehman2020rtp} and ToxiGen~\citep{hartvigsen2022toxigen} datasets. For each dataset, we sample 50k prompts for fine-tuning and another 5k for evaluation. \kw{I guess one question readers may also ask is why we want to fine-tuning LLM on RTP with constraints rather than other data, so need some sentence to motivate this choice. (as a harder setting? I guess)}\tao{I think it is common to use a toxic dataset for detoxification experiments. The general dataset is usually non-toxic and model cannot learn what is toxicity.}
    During the evaluation, we prompt the model with each data point from the evaluation set and generate 32 tokens. For the RTP dataset, we measure the average toxicity across the generations by using PerspectiveAPI. For ToxiGen, we use the pre-trained Toxigen (RoBERTa) classifier, which was released with the dataset, to calculate the percentage of generated sentences that are toxic. We test three detoxification methods, in addition to the Llama baseline:
        \begin{itemize}
            \item \textbf{Reinforcement Learning:} For each prompt in the evaluation set, we sample 32 generations. We utilize the PerspectiveAPI and ToxiGen classifier confidence scores as reward for the two test sets, respectively. We then use the policy gradient~\citep{sutton1999policy} to update the base language model.
            \item \textbf{NADO controlled decoding:} For each prompt in the two test sets, we sample 32 sentences and obtain binary labels from PerspectiveAPI and the ToxiGen classifier, respectively. When using PerspectiveAPI we set a threshold of toxicity score $>0.1$ as {\it toxic}.
            \item \textbf{Ours:} We follow the NADO-controlled decoding oracle setup. We split the 50k fine-tuning set into 5 groups. We separately run iterative sequential fine-tuning and parallel fine-tuning for 5 rounds using these groups.
        \end{itemize}
        
    \paragraph{Results} The results are shown in Tab.~\ref{tab:deto}. We observe that on both datasets our method achieves the best detoxification (given the same amount of training data). We observe that there is a significant performance improvement brought on by iterative fine-tuning when compared to NADO-controlled decoding, which shows that directly estimated the optimal distribution is not optimal. The iterative process enables the gradual push of the base model distribution towards the feasible region (Fig~\ref{fig:push}), and the estimated optimal distribution improves in its accuracy. The sequential and parallel fine-tuning results show comparable performance. Since parallel fine-tuning is more efficient, we focus on this method from this point onward.

    \tao{add one paragraph discussing preference optimization methods}
    Preference optimization \citep{rafailov2023dpo} is a popular RL method in fine-tuning LLMs. It leverages human preference between a pair of generation to achieve an alignment between model output and human. However, in our setup, the goal is to control the toxicity of the model output. The metric is clearly defined by PerspectiveAPI or ToxiGen classifier. Directly applying the toxicity value as the reward in RL is much more effective than the pairwise preference.
        
    \subsection{Balance between Utility and Detoxification}
        \begin{table*}[t]
            \centering
            \resizebox{0.8\linewidth}{!}{
            \begin{tabular}{c|c|c|c|c}
                \toprule
                \multicolumn{2}{c|}{Model} & ToxiGen & MMLU(5-shot) & Com. Reasoning (0-shot)\\
                \midrule
                \multirow{5}{*}{Llama-7B} & Baseline & 23.0 & 35.1 & 75.6 \\
                & Filtering & 21.9 & 34.6 & 75.1 \\
                & RL & 15.2 & 33.6 & 73.2 \\
                & NADO decoding & 16.8 & 31.1 & 71.4 \\
                & Ours w/o Adaptive & 13.6 & 30.4 & 71.9 \\
                & Ours w/ Adaptive & 14.2 & 33.9 & 73.6 \\
                \midrule
                \multirow{5}{*}{Falcon-7B} & Baseline & 14.0 & 27.2 & 76.1 \\
                & Filtering & 13.6 & 26.4 & 74.9\\
                & RL & 9.8 & 25.4 & 74.4 \\
                & NADO decoding & 7.3 & 23.6 & 72.5\\
                & Ours w/o Adaptive& 7.1 & 24.1 & 71.8 \\
                & Ours w/ Adaptive& 7.3 & 26.1 & 74.5 \\
                \bottomrule
            \end{tabular}}
            \caption{Benchmark performance of Llama-7B and Falcon-7B with toxicity control. The models are fine-tuned on a mixture corpus including ToxiGen and Wikitext in equal proportions. Results show that our approach achieves a better trade-off between toxicity control and benchmark performance compared to RL. Filtering is not effective in controlling toxicity. With the adaptive regularizer, LLM has a significant performance improvement on benchmarks.}
        \label{tab:multi}
        \end{table*}

        \begin{figure}[t]
            \centering
            \includegraphics[width=\linewidth]{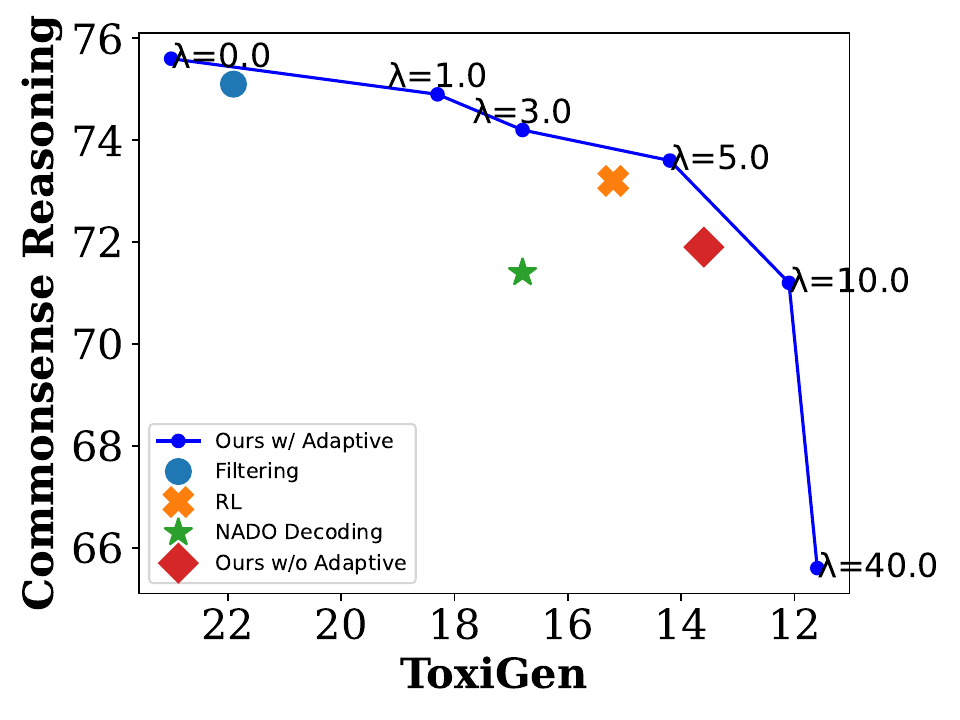}
            \caption{The trade-off curve between ToxiGen performance and Commonsense reasoning performance when fine-tuning Llama-7B model with our proposed approach with adaptive regularizer, compared to the listed baselines in Table \ref{tab:multi}. The trade-off is controlled by the coefficient $\lambda$ in Eq. \eqref{eq:adaptive}. We observe that to control the language model in the same level of toxicity, our approach, with adaptive regularizer, achieves the best commonsense reasoning performance compared to the listed methods.}
            \label{fig:tradeoff}
        \end{figure}
        
    We further study how our method can reduce toxicity generation while maintaining model utility. As RTP and ToxiGen datasets are small, fine-tuning only on them would lead to catastrophic forgetting and degradation in utility. Therefore, we fine-tune the LLM on a mix of general Wikitext corpus and toxicity corpus. We show that the proposed method achieves the best trade-off between toxicity control and maintaining performance on general utility benchmarks.

     \kw{Add a few more sentences to describe the goal/motivation of the experiment and the hypothesis we want to test here.  Also give a brief overview of the setting }\tao{Done.}
    \paragraph{Setup} We use Llama-7B~\citep{touvron2023llama} and Falcon-7B~\citep{almazrouei2023falcon} as base models, and fine-tune each of them on a mixture of ToxiGen and Wikitext~\citep{merity2016wikitext} data in equal proportions. 
    We evaluate model performance on ToxiGen toxicity, and the utility on MMLU and commensense reasoning. The details about evaluation metrics can be found in the Appendix. 
    We test 5 different methods:
        \begin{itemize}
            \item \textbf{Filtering:} We filter out all the data labeled as {\it toxic} by the ToxiGen classifier.
            \item \textbf{Reinforcement Learning:} We take the confidence score provided by the ToxiGen classifier as the reward, and apply policy-gradient to minimize the toxicity.
            \item \textbf{NADO decoding:} We train the auxiliary model on ToxiGen sampled data, and control the model generation at decoding time.
            \item \textbf{Ours (without Adaptive):} We apply parallel fine-tuning on both datasets with the auxiliary model trained on ToxiGen sampled data.
            \item \textbf{Ours (with Adaptive):} We apply an adaptive regularizer as described in Eq.~\eqref{eq:adaptive}. We use the preserving regularizer on Wikitext data, while using the toxicity control regularizer on the ToxiGen sampled data.
        \end{itemize}
        
    \paragraph{Results} The results are shown in Tab.~\ref{tab:multi}. We observe that all detoxification methods cause a performance drop on our utility metrics (i.e. MMLU and commonsense reasoning). Filtering is not effective for detoxification. In Fig. \ref{fig:tradeoff} we show the trade-off curve between ToxiGen and Commonsense reasoning tasks of our method compared to other methods. Our method with the adaptive regularizer achieves the best trade-off between toxicity control and model utility. 
        
    When used without the adaptive regularizer, our method achieves the best toxicity control. However, this comes at the cost of utility loss. This indicated that the toxicity regularizer trained on ToxiGen sampled data does not perform well on the Wikitext data. The adaptive regularizer helps preserve the model utility while fine-tuning on Wikitext data.

    We note that Falcon-7B has much lower toxicity when compared to Llama-7B. The consistent performance trends observed in both base models, demonstrate that our method is robust to different base models independent of its levels of toxicity.

        \begin{table}[t]
            \centering
            \begin{tabular}{c|c|c|c|c}
                \toprule
                Win rate & Base & Filter & RL & Ours\\
                \midrule
                Base & N/A & 44.3 & 45.1 & 51.4 \\
                Filter & 55.7 & N/A & 53.4 & 61.6 \\
                RL & 54.9 & 46.6 & N/A & 61.3 \\
                Ours & 48.6 & 38.4 & 38.7 & N/A \\
                \bottomrule
            \end{tabular}
            \caption{Pairwise comparison by OPT-30B on ToxiGen sampling data. The value shows the win rate of the method on the top row in pairwise comparison. Our model is indistinguishable from base model and outperforms Filter and RL approaching, demonstrating it retains the quality of generation. 
            \tao{add description in title.}}
        \label{tab:elo}
        \end{table}

    We further analyze model generation quality by leveraging a larger model, OPT-30B~\citep{zhang2022opt}, to do pairwise comparison on model generations for ToxiGen prompts from 4 systems: (1) the base Llama-7B model, (2) filtering, (3) RL and (4) ours with the adaptive regularizer. We do not consider NADO controlled decoding and ours without the adaptive regularizer, as they are obviously worse in terms of model quality. The results are shown in Tab.~\ref{tab:elo}. We show that OPT-30B prefers our system (with the adaptive regularizer) the best, with slight improvement over the base model.
        
    \subsection{Toxicity Classification and Generation}
        \begin{table}[t]
            \centering
            \begin{tabular}{c|c|c}
                \toprule
                Model & API Tox. & Classify ROC \\
                \midrule
                baseline & 0.315 & 0.910 \\
                SFT(LLM loss) & 0.344 & \textbf{0.966} \\
                Ours(LLM loss) & \textbf{0.288} & 0.959 \\
                \midrule
                \midrule
                SFT(classification) & 0.314 & 0.972\\
                \bottomrule
            \end{tabular}
            \caption{Jigsaw dataset performance of Llama-7B model with toxicity control. SFT with LLM loss shows a trade-off between the generation toxicity and classification performance, while our approach is capable to reduce the generation toxicity while improve toxicity classification performance.}
            \label{tab:jigsaw}
        \end{table}
    An LLM cannot avoid generating toxic outputs if it is unable to recognize toxic language. Therefore, it is essential for LLM to comprehend the characteristics of toxic content so that it can actively filter out harmful elements while maintaining the integrity of the generated output.
    However, a generic fine-tuning method often cannot improve the toxicity classification and reduce toxic generation at the same time. We design an experiment to test whether our approach can effectively enhance the LLM's ability to classify toxic content without increasing its generation of such content.
        
    \paragraph{Setup} We fine-tune the Llama-7B on the Jigsaw toxicity classification dataset~\citep{jain2022jigsaw}. We compare the performance of models fine-tuned using our controlled method to ones fine-tuned using uncontrolled fine-tuning. We use classification performance and generation toxicity (as evaluated by PerspectiveAPI) as metrics of comparison. Specifically, we compare three methods:
        \begin{itemize}
            \item \textbf{Supervised fine-tuning with LLM loss:} We concatenate each question and answer in the Jigsaw dataset, and fine-tune with a language modeling objective.
            \item \textbf{Ours with LLM loss:} We train an auxiliary model on RTP sampled data labeled by PerspectiveAPI, and fine-tune the language model same as above on Jigsaw dataset with the toxicity regularizer.
            \item \textbf{Supervised fine-tuning as classification:} We treat each question in Jigsaw as the prompt and only calculate loss on the answers. This is regarded the upper bound of performance for this task.
        \end{itemize}
        
    \paragraph{Results} The results are shown in Tab.~\ref{tab:jigsaw}. We observe that if we fine-tune the LLM on the Jigsaw dataset without toxicity control, the generation toxicity increases significantly (9.2\%, from 0.315 to 0.344). The reason is that Jigsaw consists of toxic content and fine-tuning on this shifts the model out distribution to be toxic. In comparison, when using our fine-tuning schema which leverages the toxicity regularizer, we achieve decreased toxicity. Notably our method also improves classification performance, achieving almost similar performance to uncontrolled fine-tuning, demonstrating our approach makes the model understand the toxicity rather than simply make model ignore the toxicity contents in training data.

\section{Conclusion}
We propose a novel fine-tuning approach for attribute control in LLM generations and we demonstrate its effectiveness using toxicity as our chosen attribute. While this work focuses on toxicity, our approach is general enough to accommodate other types of attributes as well. With adaptive regularizers, our method can further extend to control multiple attribute across various domains.



\kw{May add some discussion of the generalizability of the approach. We can say the approach can apply with other constraints. Although in this paper, we focus on controlling the toxicity of the models, we test on three different scenarios and demonstrate ... }\tao{how about this}
\newpage
\section*{Limitation}
In this work we assume that a decent oracle (PerspectiveAPI, ToxiGen classifier) for the attribute we would like to control is available. A low quality oracle may rely on superficial shortcut between generation and constraint labels, resulting in that the fine-tuned model captures such shortcut. Therefore, it is possible that we need to train a decent oracle before applying the proposed method. 

Although our method is general to apply different kinds of constraints since we have no assumption on the black-box oracle, in experiment we focus on detoxification. We leave the study on controlling other attributes in future work.

As a attribute control method, we note that there is a potential risk that malicious users could use this approach to 'toxify' the LLM by opposite the oracle. In addition, the generated texts may contain new toxic contents that cannot be generated in original LLM, since it may learn from the toxic fine-tuning corpus. However, on the other hand, the controlled LLM is generally less risky in generating toxic contents.

\bibliography{sample}
\bibliographystyle{acl_natbib}

\newpage
\appendix

\section{Model Architecture and Optimizer}
    In all experiments, the NADO model has the same configuration as Llama-7B model but with only 8 layers. We use AdamW as optimizer with learning rate $3e-5$ and weight decay $1e-2.$ 

    In detoxification and toxicity classification experiments, we train NADO from data sampled by base LLM. We use simple random sampling without any decoding configuration.

\section{LLM Fine-tuning}
    We fine-tune the base LLM with AdamW optimizer with learning rate $1e-5$ and weight decay $1e-2$. $\lambda=10.0$ in the multi-task scenario experiment, and $\lambda=5.0$ in the detoxification and toxicity classification experiments. 

\section{Metrics in Multitask Experiment}
    We evaluate the model performance on the following three metrics:
    \begin{compactitem}
        \item \textbf{ToxiGen (toxicity):} Same set up as the detoxification experiment in Section~\ref{sec:detox}.
        \item \textbf{MMLU (utility):} We do 5-shot evaluation on the MMLU benchmark~\citep{hendrycks2021mmlu} and report the average score.
        \item \textbf{Commonsense Reasoning (utility):} We do 0-shot evaluation on 4 commonsense reasoning benchmarks, BoolQ~\citep{clark2019boolq}, PIQA~\citep{bisk2020piqa}, HellaSwag~\citep{zellers2019hellaswag} and WinoGrande~\citep{sakaguchi2020winogrande}, and report the average score.
    \end{compactitem}

\section{Data Preprocessing}
    \paragraph{RTP and ToxiGen:} We randomly select prompts, and use the LLM to randomly sample 32 tokens in both training and evaluation. 
    
    \paragraph{Jigsaw:} The data are comment-label pairs. We templatize the data as: 
    
    The comment \textit{[comment]} is a \textit{[label name]} comment. 
    
    In evaluation, we query the model by template:
    
    Is the comment \textit{[comment]} a \textit{[label name]} comment? Answer: \textit{ [Yes / No]}

    \paragraph{MMLU and commensense reasoning:} We follow the standard o-shot and few-shot evaluation scripts.

\section{License of Datasets}
    The licenses of datasets we use in this paper list below:\\
    ToxiGen: MIT License\\
    Wikitext: CC BY-SA License and GFDL License\\
    Jigsaw: MIT License\\
    MMLU: GNU AGPL\\
    BoolQ: CC BY-SA License\\
    PIQA: Apache License\\
    HellaSwag: MIT License\\
    WinoGrande: Apache License

\end{document}